\documentclass{article}

\usepackage[dvipsnames]{xcolor}
\usepackage{microtype}
\usepackage{graphicx}
\usepackage{subfigure}
\usepackage{booktabs} 
\usepackage{caption}
\usepackage{multicol}
\usepackage{multirow}

\usepackage[accepted]{icml2024}

\usepackage{amsmath}
\usepackage{float}
\usepackage{bm}       
\usepackage{amssymb}
\usepackage{mathtools}
\usepackage{amsthm}
\usepackage{algorithm}
\renewcommand{\algorithmicrequire}{\textbf{Input:}}
\renewcommand{\algorithmicensure}{\textbf{Output:}}
\newcommand{\ctext}[2]{\textcolor{#1}{#2}}

\usepackage[capitalize,noabbrev]{cleveref}
\theoremstyle{plain}

\theoremstyle{definition}

\theoremstyle{remark}

\usepackage[textsize=tiny]{todonotes}

\usepackage{amsthm}
\newtheorem*{namedtheorem}{\textbf{Theorem 1}}

\icmltitlerunning{Graph-Mamba}

\begin{document}

\twocolumn[
\icmltitle{
\begin{tabular}{c}
\multirow{2}{*}{
\includegraphics[width=1cm, trim={2.5cm 2.5cm 2.5cm 2.5cm}, clip=True]{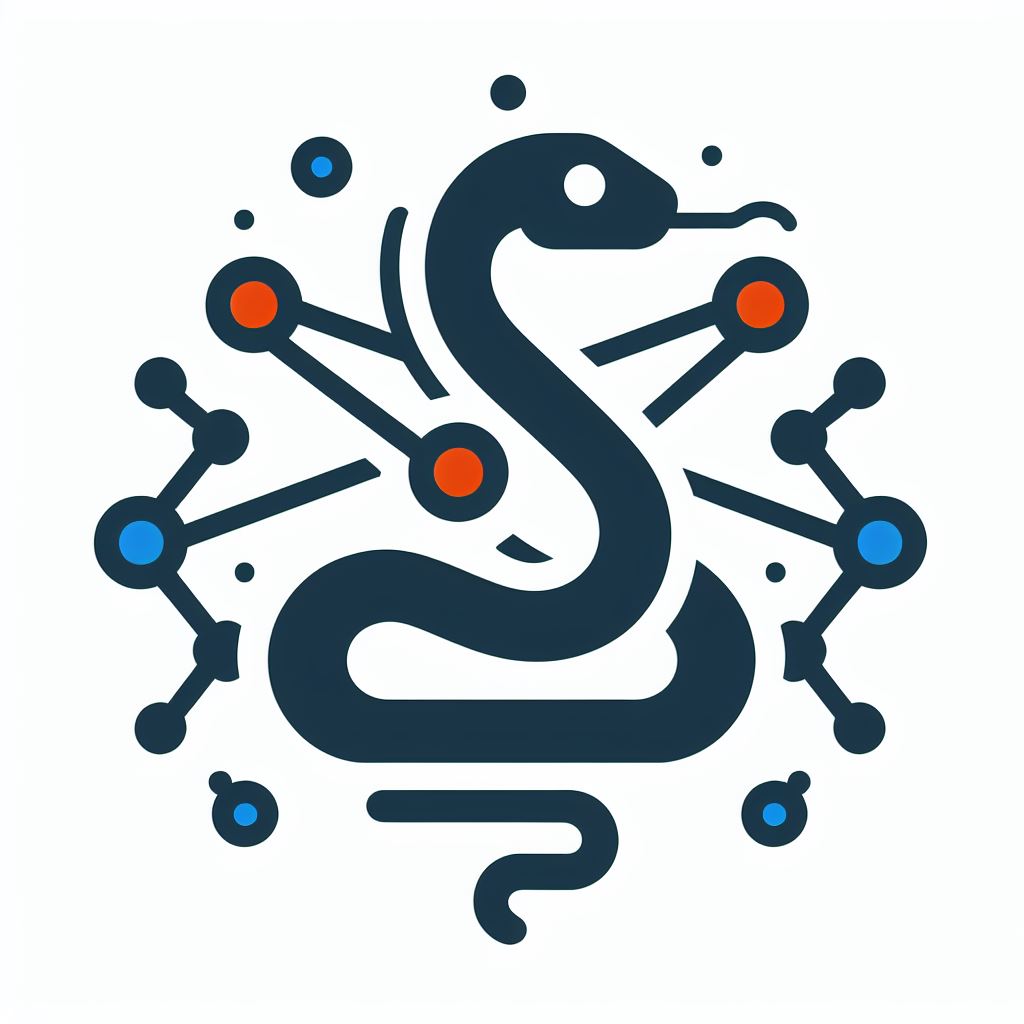}
}
\end{tabular}
Graph-Mamba: Towards Long-Range Graph Sequence Modeling with Selective State Spaces}


    \begin{icmlauthorlist}
    \icmlauthor{Chloe Wang}{cs,vector}
    \icmlauthor{Oleksii Tsepa}{cs,vector}
    \icmlauthor{Jun Ma}{vector,munc,path}
    \icmlauthor{Bo Wang}{cs,vector,munc,path,aihub}
    \end{icmlauthorlist}

    \icmlaffiliation{cs}{Department of Computer Science, University of Toronto, Toronto, Canada}
    \icmlaffiliation{vector}{Vector Institute for Artificial Intelligence, Toronto, Canada}
    \icmlaffiliation{munc}{Peter Munk Cardiac Centre, University Health Network, Toronto, Canada}
    \icmlaffiliation{path}{Department of Laboratory Medicine and Pathobiology, University of Toronto, Toronto, Canada}
    \icmlaffiliation{aihub}{AI Hub, University Health Network, Toronto, Canada}

    \icmlcorrespondingauthor{Bo Wang}{bowang@vectorinstitute.ai}
\icmlkeywords{Machine Learning, ICML}
\vskip 0.3in
]
\printAffiliationsAndNotice{} 

\begin{abstract}
Attention mechanisms have been widely used to capture long-range dependencies among nodes in Graph Transformers. Bottlenecked by the quadratic computational cost, attention mechanisms fail to scale in large graphs.
Recent improvements in computational efficiency are mainly achieved by attention sparsification with random or heuristic-based graph subsampling, which falls short in data-dependent context reasoning. 
State space models (SSMs), such as Mamba, have gained prominence for their effectiveness and efficiency in modeling long-range dependencies in sequential data. However, adapting SSMs to non-sequential graph data presents a notable challenge. In this work, we introduce Graph-Mamba, the first attempt to enhance long-range context modeling in graph networks by integrating a Mamba block with the input-dependent node selection mechanism. 
Specifically, we formulate graph-centric node prioritization and permutation strategies to enhance context-aware reasoning, leading to a substantial improvement in predictive performance. 
Extensive experiments on ten benchmark datasets demonstrate that Graph-Mamba outperforms state-of-the-art methods in long-range graph prediction tasks, with a fraction of the computational cost in both FLOPs and GPU memory consumption.
The code and models are publicly available at \url{https://github.com/bowang-lab/Graph-Mamba}.
\end{abstract}

\section{Introduction}
\label{intro}
Graph modeling has been widely used to handle complex data structures and relationships, such as social networks \cite{socialnetworks}, molecular interactions \cite{congfu}, and brain connectivity \cite{braingnn}. 
Recently, Graph Transformers have gained increasing popularity because of their strong capability in modeling long-range connections between nodes \cite{yun2019graph, dwivedi2012generalization, kreuzer2021rethinking, chen2022structure}. The typical Graph Transformer attention allows each node in the graph to interact with all the other nodes \cite{vaswani2017attention}. This serves as the complement to local message-passing approaches that primarily encode edge-based neighborhood topology \cite{kipf2016semi, xu2018powerful}. To streamline the construction of Graph Transformers, GraphGPS devised a unified framework that combines an attention module, a message passing neural network (MPNN), and structural/positional encodings (SE/PE). These components collaboratively update node and edge embeddings for downstream prediction tasks. Such decoupled pipelines offer users great flexibility to incorporate various attention modules in a plug-and-play manner~\cite{rampavsek2022recipe}. 

Although Transformers demonstrate notable enhancements of modeling capabilities, their application to long sequences is hindered by the quadratic computational cost associated with attention mechanism. This limitation has prompted further research into linear-time attention approaches. For example, BigBird \cite{zaheer2020big} and Performer \cite{choromanski2020rethinking} attempted to approximate the full attention with sparse attention or lower-dimensional matrices. However, designed for sequential inputs, BigBird does not generalize well to non-sequential inputs such as graphs, leading to performance deterioration in GraphGPS \cite{shirzad2023exphormer}. Exphormer tailored such attention sparsification principles to graph input, by incorporating local connectivity defined by edges as local attention \cite{shirzad2023exphormer}. These adaptations led to improved performance comparable to full graph attention.

However, approximating the full attention, or encoding all contexts, may not be ideal for long-range dependencies. In empirical observations, many sequence models do not improve with increasing context length \cite{gu2023mamba}. Mamba, a selective state space model (SSM), was proposed to solve data-dependent context compression in sequence modeling \cite{gu2023mamba}. Instead of attention computation, Mamba inherits the construct of state space models that encode context using hidden states during recurrent scans. The selection mechanism allows control over which part of the input can flow into the hidden states, as part of the context that influence subsequent embedding updates. In graph modeling, this can be viewed as a data-dependent node selection process. By filtering relevant nodes at each step of recurrence and ``attending'' only to this selected context, Mamba helps achieve the same objectives as attention sparsification, serving as an alternative to random subsampling. Moreover, the Mamba module is optimized for linear time complexity and reduced memory, offering improved efficiency for large-graph training tasks. However, challenges present in effectively adapting Mamba designed for sequence modeling to non-sequential graph input.

\textbf{Contributions.} Motivated by the exceptional long-sequence modeling ability of Mamba, we propose Graph-Mamba to alleviate the high computational cost associated with Graph Transformers, as a data-dependent alternative for attention sparsification. In particular, this work presents the following key contributions:

\begin{itemize}
\item \textbf{Innovative graph network design:} Graph-Mamba represents a new type of graph network pioneering integration with selective state space models, which performs input-dependent node filtering and adaptive context selection. The selection mechanism captures long-range dependencies and improves on the existing subsampling-based attention sparsification techniques.


\item \textbf{Adaptation of SSMs for non-sequential graph data:} We designed an elegant way to extend state space models to handle non-sequential graph data. Specifically, we introduced a node prioritization technique to prioritize important nodes for more access to context, and employed a permutation-based training recipe to minimize sequence-related biases.

\item \textbf{Superior performance and efficiency:} Comprehensive experiments on ten public datasets demonstrate that Graph-Mamba not only outperforms baselines but also achieves linear-time computational complexity. Remarkably, Graph-Mamba reduces GPU memory consumption by up to 74\% on large graphs, highlighting its efficiency in long-range graph datasets.

\end{itemize}

\section{Related Work}
\label{relatedworks}

\subsection{Graph Neural Networks}
Graph Neural Networks (GNNs) leverage message passing as a key mechanism for graph modeling, enabling nodes to communicate and iteratively aggregate information from their neighbors. Graph Convolutional Networks (GCN) \cite{kipf2017semisupervised, defferrard2017convolutional} pioneered GNNs, influencing subsequent works like GraphSage \cite{hamilton2018inductive}, GIN \cite{xu2018powerful}, GAT \cite{veličković2018graph}, and GatedGCN \cite{bresson2018residual}. Despite the significance in aggregating node features based on graph topology, MPNNs have limited expressive power upper bounded by the 1-dimensional Weisfeiler-Lehman (1-WL) isomorphism test \cite{xu2018powerful}. Additionally, aggregated node features are prone to over-smoothing in the local neighborhood \cite{alon2021bottleneck, topping2022understanding}.



\subsection{Graph Transformers}

Transformers with attention mechanism have achieved unprecedented success in various domains, including natural language processing (NLP) \cite{vaswani2017attention, kalyan2021ammus} and computer vision \cite{pmlr-v139-d-ascoli21a, guo2022cmt, dosovitskiy2021image}. Graph Transformers typically compute full attention, allowing each node to attend to all others, regardless of the edge connectivity. This enables Graph Transformers to effectively capture long-range dependencies, avoiding over-aggregation in local neighborhoods like MPNNs. However, full attention, with its $O(N^2)$ complexity, fails to scale in large graphs.


Analogous to positional embeddings in transformers for NLP, the first Graph Transformer \cite{dwivedi2021generalization} introduced graph Laplacian eigenvectors as node PE. Subsequently, SAN \cite{san} incorporated invariant PE aggregation, integrating conditional attention for both real and virtual graph edges. Concurrently, Graphormer \cite{graphormer} integrated relative PE into the attention mechanism using centrality and spatial encodings. GraphiT \cite{mialon2021graphit} utilized relative PE based on diffusion kernels to simulate the attention mechanism. Lastly, GraphTrans \cite{graphtrans} proposed a two-stage architecture, employing a graph transformer on local embeddings derived from MPNNs.


\subsection{GraphGPS}
GraphGPS \cite{rampavsek2022recipe} employed a modular framework that integrates SE, PE, MPNN, and a graph transformer. Users have the flexibility to choose the methods for each component within this framework. Given an input graph, GraphGPS computes SE and PE, concatenates them with node and edge embeddings, and passes these embeddings into the GPS layers. In the GPS layers, a graph transformer and MPNN collaboratively update the node and edge embeddings. The GraphGPS framework allows the replacement of fully-connected Transformer attention with its sparse alternatives, resulting in $O(N+E)$ complexity. 

\subsection{Sparse Graph Attention}
BigBird \cite{zaheer2020big} and Performer \cite{choromanski2020rethinking} are the two sparse attention methods supported by GraphGPS. Performer improved computation efficiency by using lower-dimensional positive orthogonal random features to approximate softmax kernels in regular attention. BigBird employed graph subsampling and sequence heuristics to approximate full attention, combining randomly subsampled attention, local attention among adjacent tokens, and global attention with global tokens \cite{zaheer2020big}. Randomly subsampled graphs, or expanders, are known to approximate the spectral properties of full graphs \cite{spielman2011spectral, yun2020n}. However, BigBird's local attention is specifically designed for sequence input with a sliding window on adjacent tokens, making it unsuitable for modeling graph input. Exphormer \cite{shirzad2023exphormer} proposed a graph adaptation of BigBird by incorporating local neighborhood attention among neighbors defined by edges, and global attention connecting virtual nodes to all nodes, to expanders. These adaptations further improved model performance while benefiting from the linear complexity of sparse attention. However, the random node subsampling process suggests potential room for improvement. Incorporating methods that allow informed context selection could serve as further enhancement.




\subsection{State Space Models}
General state space models involve recurrent updates over a sequence through hidden states. Implementations range from hidden Markov models to recurrent neural networks (RNNs) in deep learning. Utilizing a recurrent scan, SSM stores context in its hidden states, and updates the output by combining these hidden states with input. Structured state space models (S4) enhance computational efficiency with reparameterization \cite{gu2021efficiently}, offering an efficient alternative to attention computation. Recent S4 variants for linear-time attention include H3 \cite{fu2022hungry}, Gated State Space \cite{mehta2022long}, Hyena \cite{nguyen2023hyenadna}, and RWKV \cite{peng2023rwkv}. Mamba further introduces a data-dependent selection mechanism to S4 to capture long-range context with increasing sequence length \cite{gu2023mamba}. Notably, Mamba demonstrates linear-time efficiency in long-sequence modeling, and outperforms Transformers on various benchmarks. Mamba has also been successfully adapted for non-sequential input such as images on segmentation tasks to enhance long-range dependencies \cite{ma2024u, zhu2024vision,liu2024vmamba}.

\section{Graph-Mamba}
\label{model}

\begin{figure*}[ht]
\begin{center}
\centerline{\includegraphics[width=\textwidth, trim={0 3cm 6cm 1cm}, clip]{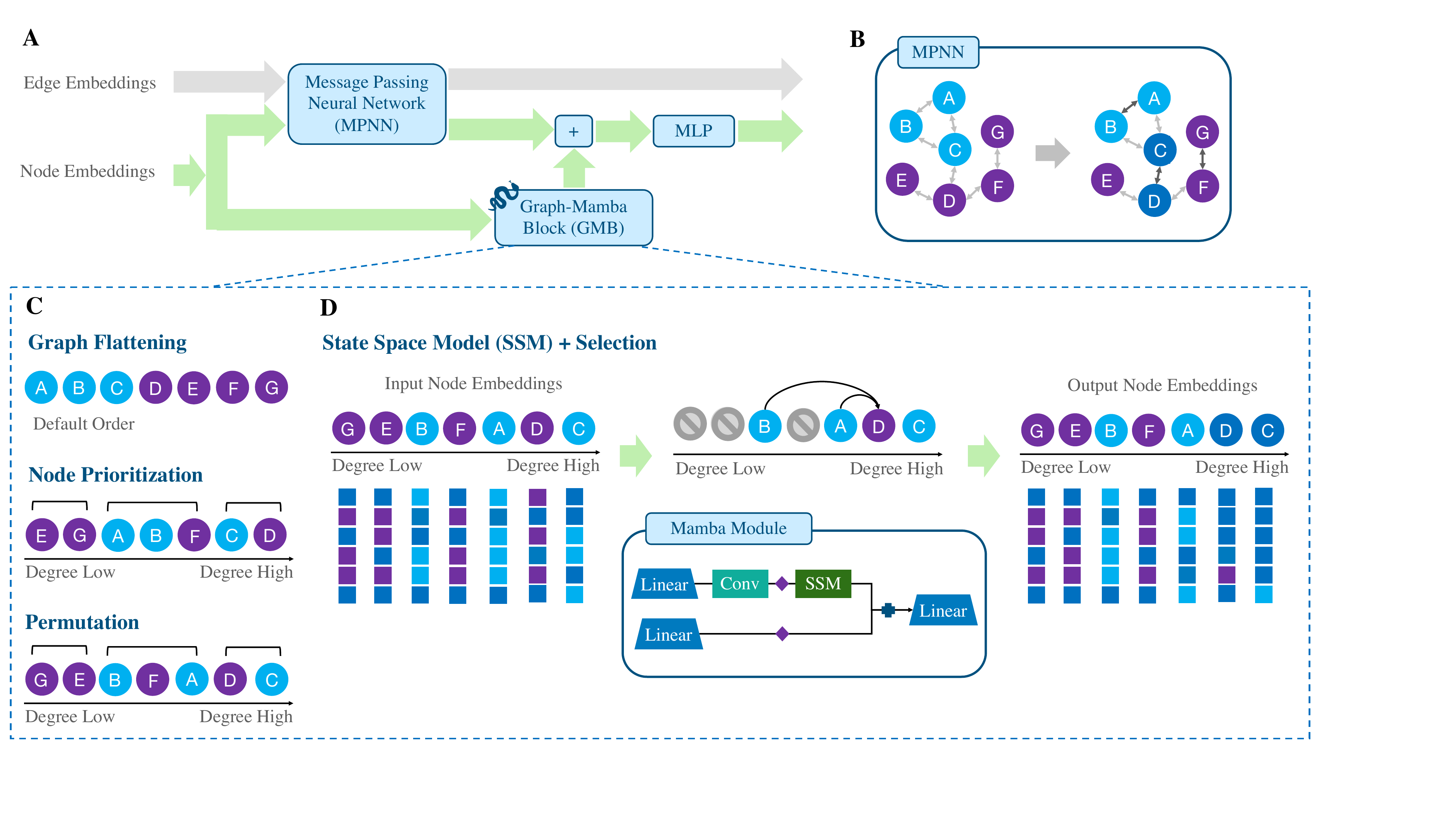}}
\caption{\textbf{Overview of Graph-Mamba architecture}, by incorporating GMB to replace the attention module in the GraphGPS framework. A) The GMB layer, an adaptation of GPS layer that combines an edge-based MPNN and a node-focused GMB to output updated node and edge embeddings. B) Graph-Mamba employs the GatedGCN model as default for MPNN. C) GMB's specialized training recipe with node prioritization  and permutation techniques that perform informed graph sparsification. D) The selection mechanism with Mamba module that facilitates input-dependent context filtering.}
\label{fig:fig1}
\end{center}
\vskip -0.2in
\end{figure*}

Graph-Mamba employs a selective SSM to achieve input-dependent graph sparsification. In particular, we have designed a Graph-Mamba block (GMB) and incorporated it into the popular GraphGPS framework, enabling fair comparisons with other graph attention implementations. GMB leverages the recurrent scan in sequence modeling with a selection mechanism to achieve two levels of graph sparsification. The first level involves the selection mechanism in Mamba module, which effectively filters relevant information within the long-range context.  The second level is achieved through the proposed node prioritization approach, allowing important nodes in the graph to access more context. 
Consequently, these sequence modeling features present a promising avenue of combining data-dependent and heuristic-informed selection for graph sparsification. Morever, Graph-Mamba implementation using the Mamba module ensures linear-time complexity, as an efficient alternative to dense graph attention.

To contextualize SSMs in graph modeling, we first review SSMs followed by the selection mechanism in Section \ref{mamba} and \ref{memory}. Next, we introduce the Graph-Mamba implementation in Section \ref{gps}, and detail GMB's specialized graph adaptation techniques in Section \ref{sorting} and \ref{trainingrecipe}. Finally, we discuss the computational efficiency of GMB in Section \ref{mamba-runtime}.


\subsection{Structured state space models for sequence modeling}\label{mamba}
SSM is a type of sequence model that defines a linear Ordinary Differential Equation (ODE) to map input sequence $x(t) \in \mathbb{R}^N$  to output sequence $y(t) \in  \mathbb{R}^N$ by a latent state $h(t)\in \mathbb{R}^N$:
\begin{equation}
\begin{aligned}
    h'(t) &= \bm{A}h(t) + \bm{B}x(t), \\
    y(t) &= \bm{C}h(t),
\end{aligned}
\end{equation}\label{eq:ssm-con}
where $\bm{A}\in \mathbb{R}^{N\times N}$ and $\bm{B}, \bm{C}\in \mathbb{R}^N$ denote the state matrix, input matrix, and output matrix, respectively. 
To obtain the output sequence $y(t)$ at time $t$, we need to find $h(t)$ which is difficult to solve analytically. In contrast, real-world data is usually discrete rather than continuous. As an alternative, we discretize the system Equation (1) as follows:
\begin{equation}\label{SSM}
\begin{split}
    h_{t} & = \bm{\bar{A}}h_{t-1} + \bm{\bar{B}}x_{t}, \\
    y_{t} & = \bm{C} h_{t},
\end{split}
\end{equation}
where $\bm{\bar{A}}:=\exp(\Delta \cdot \bm{A})$ and $\bm{\bar{B}}:=(\Delta \cdot \bm{A})^{-1}(\exp(\Delta\cdot \bm{A})-I)\cdot \Delta \bm{B}$ are the discretized state parameters and $\Delta$ is the discretization step size. 
SSMs have rich theoretical properties but suffer from high computational cost and numerical instability. Structured state space sequence models (S4) addressed these limitations by imposing structure on the state matrix $\bm{A}$ based on HIPPO matrices, which significantly improved the performance and efficiency. In particular, S4 surpassed Transformers by a large margin on the Long Range Arena benchmark, which requires effective modeling of the long-range dependencies \cite{gu2021efficiently}.





\subsection{Graph-dependent selection mechanism}\label{memory}
S4 has demonstrated better suitability for modeling long sequences, but underperforms when content-aware reasoning is needed, attributed to its time-invariant nature. More specifically, $\bm{A}, \bm{B}$, and $ \bm{C}$ are the same for all input tokens in a sequence. 
Mamba \cite{gu2023mamba} addressed this issue by introducing the selection mechanism, allowing the model to adaptively select relevant information from the context. This can be achieved by simply making the SSM parameters $\bm{B}, \bm{C}$, and $\Delta$ as functions of the input $x$.
Furthermore, a GPU-friendly implementation is designed for efficient computing of the selection mechanism, which significantly reduces the number of memory IOs and avoids saving the intermediate states. 



We use the reparameterized discretization step size $\Delta$ as an example to illustrate the intuition behind Mamba's selection mechanism. Revisiting \citet{gu2023mamba}'s main Theorem, $\Delta_{t}$ assumes a generalized role related to gating mechanism in RNN to facilitate input-dependent selection, where $g_{t} = \sigma(Linear(x_{t}))$ and $h_{t} = (1-g_{t})h_{t-1} + g_{t}x_{t}$, as detailed in Appendix \ref{proof-theorem}. Intuitively, the current input $x_{t}$ is able to control the balance between current input and previous context $h_{t-1}$ with $g_{t}$ when updating the hidden state $h_{t}$. This is achieved by parameterizing $\Delta$ as a function of input in the discretization step to obtain data-dependent $\bm{\bar{A}}$ and $\bm{\bar{B}}$, which acts as the main selection mechanism similar to gating in RNN. Additionally, the projection matrices $\bm{B}$ and $\bm{C}$
are parameterized as linear projections of the input $x$, to further control how much $x_{t}$ updates the hidden states and how much $h_{t}$ influences the output $y_{t}$

In graph learning, with nodes as input sequence, the selection mechanism allows the hidden states to update based on relevant nodes from prior sequence, gated by the current input node, and subsequently influencing the current node's output embeddings. $g_{t}$ ranges between 0 and 1, allowing the model to filter out irrelevant context entirely when needed. The ability to select and reset enables Mamba to distill relevant dependencies given long-range context, while minimizing the influence of unimportant nodes at each step of recurrence. It hence offers a context-aware alternative for sparsifying graph attention, by retaining relevant dependencies only in the long input sequences.

\subsection{Graph-Mamba workflow}\label{gps}
\begin{algorithm}\label{algo:algo1}
\caption{GMB Forward Pass}\label{alg:cap}
\begin{algorithmic}[1]
\REQUIRE Node embeddings $\bm{X} \in \mathbb{R}^{L\times D}$; Node heuristic $\bm{H} \in \mathbb{R}^{L\times 1}$.
\ENSURE Updated node embeddings $\bm{X'} \in \mathbb{R}^{L\times D}$.
\STATE$\bm{H'} = \bm{H} + Noise(0, 1)$ \COMMENT{Add noise to heuristic for node shuffling}
\STATE$\bm{I_{sorted}} \gets Argsort(\bm{H'})$
\STATE$\bm{I_{reverse}} \gets Argsort(\bm{I_{sorted}})$
\STATE $\bm{X_{sorted}} : (L,D) \gets \bm{X[I_{sorted}]}$ \COMMENT{Sort nodes by node heuristic}
\STATE $\bm{X_{norm}} : (L,D) \gets LayerNorm(\bm{X_{sorted}})$ \COMMENT{Input to SSM + Selection}
\STATE $\bm{x} : (L,D') \gets Linear^{0}(\bm{X_{norm}})$
\STATE $\bm{x'} : (L,D') \gets SiLU(Linear^{1}(\bm{X_{norm}}))$ 
\STATE $\bm{x^{SSM}} (L,D') \gets SiLU(Conv1d(\bm{x}))$
\STATE $\bm{B} : (L, N) \gets Linear^{B}(\bm{x^{SSM}})$
\STATE $\bm{C} : (L, N) \gets Linear^{C}(\bm{x^{SSM}})$
\STATE $\bm{\Delta} : (L, D') \gets softplus(Linear^{\Delta}(\bm{x^{SSM}}))$
\STATE $\bm{\bar{A}} : (L, D', N) \gets discretize^{A}(\bm{\Delta}, \bm{A})$
\STATE $\bm{\bar{B}} : (L, D', N) \gets discretize^{B}(\bm{\Delta}, \bm{A}, \bm{B})$
\STATE $\bm{y} : (L, D') \gets SSM(\bm{\bar{A}}, \bm{\bar{B}}, \bm{C})(\bm{x^{SSM}})$
\STATE $\bm{y'} : (L, D') \gets \bm{y} \odot \bm{x'}$
\STATE $\bm{X_{sorted}'} : (L, D) \gets Linear^{2}(\bm{y'})$
\STATE $\bm{X'} : (L, D) \gets \bm{X_{sorted}'}[\bm{I_{reverse}}]$ \COMMENT{Reverse sort output}
\end{algorithmic}
\end{algorithm}

Graph-Mamba incorporates Mamba's selection mechanism from Section \ref{memory} into the GraphGPS framework. Figure \ref{fig:fig1} A illustrates Graph-Mamba's adaptation of the GPS layer, where the attention module is replaced by GMB, denoted as a GMB layer. We used the GatedGCN model as the default for MPNN for local context selection, as shown in Figure \ref{fig:fig1} B. The GatedGCN model aggregates information from neighboring nodes defined by edge connections, and employs a gating mechanism to decide on how much of that information to incorporate, inspired by RNNs. GatedGCN and GMB collectively contribute to the overarching theme of recurrence-based context selection in Graph-Mamba, facilitating node filtering within the local neighborhood and among the global connections. The graph feature computation with SE and PE prior to the GMB layers remain consistent. The GMB layers thus receive the SE/PE-aware node and edge embeddings as input. 

A Graph-Mamba framework consists of $K$ stacked GMB layers. Algorithm 1 defines the GMB function (more explanations in Sections 3.4-3.6), and Algorithm 2 illustrates the forward pass through $K$ GMB layers. In Algorithm 2, each GMB layer performs two round of embedding updates using MPNN and GMB, given an input graph of $L$ nodes, $E$ edges, and embedding size $D$. Specifically, an MPNN updates both node and edge embeddings (line 2), while GMB updates the node embeddings only (line 3). The updated node embeddings from an MPNN ($\bm{X}^{k+1}_{M}$) and GMB ($\bm{X}^{k+1}_{GMB}$) are combined through an MLP layer to produce the output node embeddings (line 6). Using the output from the previous layer as the input for the next layer, this process iterates through $L$ GMB layers to obtain the final output node embeddings, which are subsequently used for downstream tasks.

\subsection{Node prioritization strategy for non-sequential graph input}\label{sorting}

\begin{center}
\centering
    \begin{table*}[t]
    \centering
    \captionsetup{justification=centering}
    \caption{\textbf{Benchmark of Graph-Mamba on Long-Range Graph Datasets} with existing methods. These five datasets feature large input graphs with 150 to 1,400 nodes. Best results are colored in \ctext{ForestGreen}{\textbf{first}}, \ctext{BurntOrange}{\textbf{second}}, \ctext{Periwinkle}{\textbf{third}}.}
     \begin{tabular}{l c c c c c cccl } \hline \hline
     \textbf{Model} & \textbf{Peptides-Func} & \textbf{Peptides-Struct} & \textbf{PascalVOC-SP} & \textbf{COCO-SP} & \textbf{MALNET-TINY}  \\
      & AP $\uparrow$ & MAE $\downarrow$ & F1 score $\uparrow$ & F1 score $\uparrow$ & Accuracy $\uparrow$  \\
     \hline
     GCN  & 0.5930$\pm$0.0023 & 0.3496$\pm$0.0013 & 0.1268$\pm$0.0060 & 0.0841 $\pm$ 0.0010  & 0.8100  \\
     GIN  & 0.5498$\pm$0.0079 & 0.3547$\pm$0.0045 & 0.1265$\pm$0.0076 & 0.1339$\pm$0.0044  &  0.8898$\pm$0.0055\\
     GatedGCN  & 0.5864$\pm$0.0077 & 0.3420$\pm$0.0013 & 0.2873$\pm$0.0219 & 0.2641$\pm$0.0045 &  0.9223$\pm$0.0065 \\
     \hline
    GPS+Transformer  & \ctext{BurntOrange}{\textbf{0.6575$\pm$0.0049}} & \ctext{BurntOrange}{\textbf{0.2510$\pm$0.0015}} & \ctext{Periwinkle}{\textbf{0.3689$\pm$0.0131}} & \ctext{BurntOrange}{\textbf{0.3774$\pm$0.0150}} & OOM (bs=8) \\
    GPS+Performer  & \ctext{Periwinkle}{\textbf{0.6475$\pm$0.0056}} & 0.2558$\pm$0.0012 & \ctext{BurntOrange}{\textbf{0.3724$\pm$0.0131}} & \ctext{Periwinkle}{\textbf{0.3761$\pm$0.0101}} & \ctext{Periwinkle}{\textbf{0.9264$\pm$0.0078}} \\
    GPS+BigBird  & 0.5854$\pm$0.0079 & 0.2842$\pm$0.0130 &   0.2762$\pm$0.0069& 0.2622$\pm$0.0008& 0.9234$\pm$0.0034 \\
    Exphormer  & 0.6258$\pm$0.0092 & \ctext{Periwinkle}{\textbf{0.2512$\pm$0.0025}} & 0.3446$\pm$0.0064 & 0.3430$\pm$0.0108 & \ctext{ForestGreen}{\textbf{0.9422$\pm$0.0024}} \\
    \hline
    Graph-Mamba  & \ctext{ForestGreen}{\textbf{0.6739$\pm$0.0087}} & \ctext{ForestGreen}{\textbf{0.2478$\pm$0.0016}} &  \ctext{ForestGreen}{\textbf{0.4191$\pm$0.0126}} & \ctext{ForestGreen}{\textbf{0.3960$\pm$0.0175}} &  \ctext{BurntOrange}{\textbf{0.9340$\pm$
    0.0027}} \\
     \hline \hline
    \end{tabular}
    \vspace{0.1cm}
    \label{tab:lrgb}
    \end{table*}
\vskip -0.3in
\end{center}

\begin{algorithm}\label{algo:algo2}
\caption{Graph-Mamba with K GMB Layers}\label{alg:cap2}
\begin{algorithmic}[1]
\REQUIRE Node embeddings $\bm{X}^{0} \in \mathbb{R}^{L\times D}$;  Edge embeddings $\bm{E}^{0} \in \mathbb{R}^{E\times D}$; Adjacency matrix $\bm{A} \in \mathbb{R}^{L\times L}$.
\ENSURE Returns $\bm{X}^{K} \in \mathbb{R}^{L\times D}$, and  $\bm{E}^{K} \in \mathbb{R}^{E\times D}$, used for downstream prediction tasks.

\FOR {$k$ = 0, 1, · · · , K-1}
\STATE $\hat{\bm{X}}^{k+1}_{M}, \bm{E}^{k+1} \gets MPNN^{k}(\bm{X}^{k}, \bm{E}^{k}, \bm{A})$
\STATE$\hat{\bm{X}}^{k+1}_{GMB} \gets GMB^{k}(\bm{X}^{k})$
\STATE $\bm{X}^{k+1}_{M} \gets Dropout(\hat{\bm{X}}^{k+1}_{M} + \bm{X}^{k})$
\STATE $\bm{X}^{k+1}_{GMB} \gets Dropout(\hat{\bm{X}}^{k+1}_{GMB} + \bm{X}^{k})$
\STATE $\bm{X}^{k+1} \gets MLP^{k}(\bm{X}^{k+1}_{M}+\bm{X}^{k+1}_{GMB})$
\ENDFOR
\STATE {return $\bm{X}^{K}, \bm{E}^{K}$}
\end{algorithmic}
\end{algorithm}

A major challenge of adapting sequence models such as Mamba to graphs stems from the unidirectionality of recurrent scan and update. In dense attention, all nodes attend to one another. However, due to the recurrent nature of the sequence modeling, in Mamba, each node gets updated based on nodes that come before them from the hidden states, not vice versa. For example, in an input sequence of length $L$, the last node has access to hidden states that incorporate most context including all prior nodes 0 to $L-2$. In contrast, node 1 only has access to limited context via hidden states that encode node 0 only.
This restricted information flow removes connections between nodes based on its position in the sequence, allowing GMB to prioritize specific nodes of higher importance at the end of the sequence for informed sparsification. 

To achieve informed sparsification in GMB, we explored an input node prioritization strategy by node heuristics that are proxy of node importance, as illustrated in Figure \ref{fig:fig1} C. When we first flatten a graph into a sequence, the nodes do not assume any particular order. The input nodes are then sorted in ascending order by node heuristic such as node degree. The intuition behind is that more important nodes should have access to more context (i.e., a longer history of prior nodes), and therefore to be placed at the end of the sequence. In in Algorithm 1, lines 1-4 illustrate the sequence sorting procedure for an input graph of $L$ nodes, where node heuristic $\bm{H'}$ determines the node order in the flattended sequence. Lines 5-16 compute the selective SSM using Mamba, explained in more details in subsequent Section \ref{mamba-runtime}. In line 17, the SSM output is reverse sorted to return updated $\bm{X'}$ in the original order. More details about other choices of node heurstics and rationale behind node prioritization are summarized in Appendix \ref{supp:seqconstruction}.

\subsection{Permutation-based training and inference recipe}\label{trainingrecipe}

Following the input node prioritization strategy, Graph-Mamba uses a permutation-focused training and inference recipe to promote permutation invariance, as illustrated in Figure \ref{fig:fig1} C. Intuitively, when ordering the nodes by heuristics such as node degree, nodes within the same degree are deemed equally important in the graph. Therefore, nodes of the same degree are randomly shuffled during training to minimize bias towards any particular order. Line 1 in Algorithm 1 illustrates the permutation implementation. Specifically, random noise $\in [0, 1)$ is added to node heuristic $\bm{H}$, and the jittered $\bm{H'}$ determines the input node order.

In the training stage of Graph-Mamba, GMB is called once to output updated node embeddings from a random permutation of input node sequence. At inference time, 
the $m$ GMB outputs $\hat{\bm{X}}^{k+1}_{GMB}$ are averaged and passed on to subsequent computation. The $m$-fold average at inference time aims to provide stability, and makes the output node embeddings invariant to the permutations applied.



\begin{figure*}[ht!]
\begin{center}
\centerline{\includegraphics[width=0.9\textwidth, trim={0cm 10.5cm 8.5cm 0.1cm}, clip]{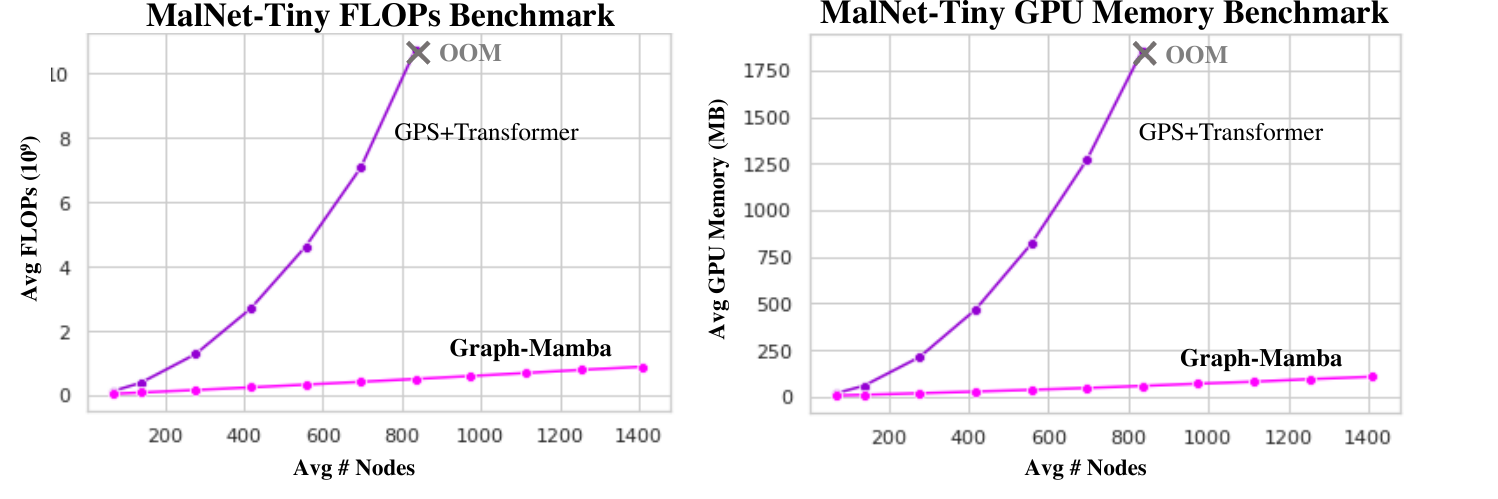}}
\caption{\textbf{FLOPs and Memory Benchmark of Graph-Mamba with GPS+Transformer} on the MalNet-Tiny dataset, subsampled at various ratios.
}
\vskip -0.2in
\label{fig:fig2}
\end{center}
\end{figure*}

\subsection{GMB with improved computation efficiency}\label{mamba-runtime}

GMB's selection mechanism is illustrated in Figure \ref{fig:fig1} D, with corresponding Mamba implementation detailed in Algorithm 1 lines 5-16. The sorted node sequence $\bm{X_{sorted}}$ consists of $L$ nodes, each with a node embedding of size $D$. The Mamba computation consists of linear projection of normalized input to $D'$ dimensions (line 5, 6), followed by 1-D convolution and SiLU activation (line 8), and SSM computation (lines 9-14). The SSM output $y$ is gated by a projection of the original input (line 7, 15), before the final projection back the original size as output (line 16). 

With the data-dependent selection mechanism, the $L$-fold expansion in parameters in $\bm{\bar{A}}$, $\bm{\bar{B}}$, and $\bm{C}$ would lead to increased computational cost in SSM. Mamba implements an efficient hardware-aware algorithm that leverages the hierarchy in GPU memory to alleviate this overhead. Specifically, with input batch size $B$, Mamba reads the $O(BLD'+ND')$ of input $\bm{{A}}$, ${\bm{B}}$, $\bm{C}$, and $\Delta$ from HBM, computes the intermediate states of size $O(BLD'N)$ in SRAM and writes the final output of size of $O(BLD')$ to HBM, thus reducing IOs by a factor of $N$. Not storing the intermediate states also lowers memory consumption, where intermediates states are recomputed for gradient calculation in the backward pass. 
With the GPU-aware implementation of Mamba, GMB achieves linear time complexity ($O(L)$) to input sequence length, which is significantly faster than the dense attention computation in transformers with quadratic time complexity ($O(L^2)$).

\section{Experiments}
\label{experiments}
\subsection{Benchmark on graph-based prediction tasks}

We benchmarked Graph-Mamba on ten datasets from the Long Range Graph Benchmark (LRGB) \cite{dwivedi2022long} and GNN Benchmark \cite{dwivedi2023benchmarking} as a comprehensive evaluation. These benchmarks evaluate model performance on various graph-based prediction tasks, including graph, node, and link-level classification and regression. For each dataset, we reported the test metric across multiple runs to ensure robustness. The dataset and task descriptions are summarized in Appendix \ref{app:dataset}. Details about experiment setup are summarized in Appendix  \ref{app:hp} and \ref{app:exp}.

We focused the comparison on the choice of attention modules within the GraphGPS framework. Specifically, we evaluated Graph-Mamba's performance against GraphGPS with dense attention (Transformer) and various implementations of sparse attention (i.e., Exphormer, Performer, and BigBird) \cite{rampavsek2022recipe, shirzad2023exphormer, choromanski2020rethinking, zaheer2020big}.

Table \ref{tab:lrgb} highlights Graph-Mamba's superior performance in capturing long-range dependencies from the top five datasets with the largest input lengths, ranging from 150 to 1,400 nodes per graph respectively. In four out of five datasets, Graph-Mamba offered considerable improvement (up to 5\%) to the other sparse attention methods. Graph-Mamba also compared favorably to Transformer with dense attention, which underscores the importance of context-aware node selection in graphs with long-range dependencies. On the other five datasets with small to medium-sized graphs, Graph-Mamba further demonstrated its robustness by showcasing comparable performance to the state-of-the-art sparse and dense attention methods, summarized in Appendix Table \ref{tab:gnnbenchmark}. These results endorsed Graph-Mamba's ability to capture long-range context with the input-dependent node selection mechanism, while generalizing well to common graph-based prediction tasks.

\subsection{FLOPs and memory consumption}

\begin{figure*}[ht!]
\begin{center}
\centerline{\includegraphics[width=\textwidth, trim={0cm 8cm 3cm 0.1cm}, clip]{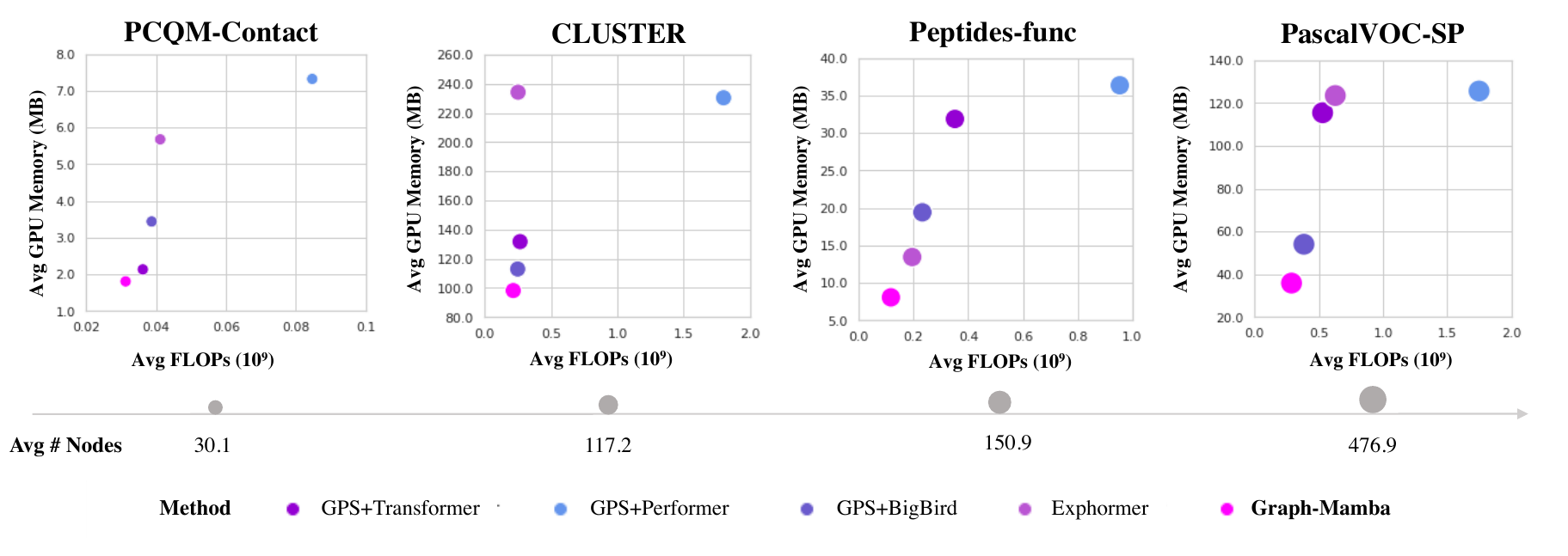}}
\caption{\textbf{FLOPs and Memory Benchmark of Graph-Mamba} with existing methods, with emphasis on \textbf{sparse attention variants}, on four datasets of increasing input size.
}
\vskip -0.2in
\label{fig:fig3}
\end{center}
\end{figure*}

\begin{table*}[ht!]
\caption{\textbf{Ablation Study of Permutation and Node Prioritization Strategies} on the Peptides-Func and PascalVOC-SP datasets.} 
\centering
 \begin{tabular}{ l | l | c c } 
 \hline \hline
  \textbf{Permutation} & \textbf{Node Prioritization} & \textbf{Peptides-Func} & \textbf{PascalVOC-SP} \\ 
  & & AP $\uparrow$ &  F1 score $\uparrow$ \\
 \hline
 \hline
   - &  - & 0.6581 &  0.3105\\ 
\hline
   Node Level & - & 0.6821 & 0.4193 \\ 
\hline
   Node Level & Degree & \textbf{0.6834} & \textbf{0.4314}
\\
 \hline \hline
\end{tabular}
\vspace{0.1cm}
\label{tab:seq_sort}
\end{table*}

Graph-Mamba offers significant improvement in efficiency in addition to performance gain. We benchmarked Graph-Mamba's Floating Point Operations (FLOPs) and memory consumption during training stage against existing methods on five datasets, as detailed in Appendix \ref{app:exp}. 

Figure \ref{fig:fig2} illustrates the computational cost on the MalNet-Tiny dataset with an average of 1,410.3 nodes, subsampled at increasing ratios. Graph-Mamba demonstrates linear complexity in both FLOPs and memory with respect to input length, whereas GPS-Transformer's cost grows quadratically. GPS-Transformer encounters out-of-memory issues with input sizes below 700 nodes at a batch size of 16, impeding efficient model training. In contrast, Graph-Mamba supports the training of full graphs with twice the number of nodes, at batch size up to 256. 

In Figure \ref{fig:fig3}, we further compared Graph-Mamba with the sparse attention variants in GPS. For each benchmark dataset, the x-axis represents the average number of FLOPs per training example, while the y-axis showcases the average GPU consumption.
Graph-Mamba consistently occupies the lower left corner, indicating the fewest FLOPs and least memory usage across all datasets. Specifically, in the Peptides-func dataset, Graph-Mamba achieves a 74\% reduction in memory usage and a 66\% reduction in FLOPs compared to Transformer. Moreover, Graph-Mamba demonstrates a 40\% decrease in both FLOPs and memory usage against the state-of-the-art sparse graph attention implementation, Exphormer. These results highlight the computational efficiency of Graph-Mamba, opening up the potential of efficient training on larger graphs.

\subsection{Ablation of Graph-Mamba's training and inference recipes}

We demonstrate the effectiveness of the proposed training and inference recipe for Graph-Mamba with an ablation study. Table \ref{tab:seq_sort} showcases the predictive performance of Graph-Mamba from three training and inference settings on two datasets Peptides-Func and PascalVOC-SP. The baseline training procedure did not use any node prioritization or permutation techniques. The permutation-only setting introduced random permutation at the node-level during training time. At inference time, the output was obtained by averaging five runs of corresponding permutation. The node prioritization setting injected node degree as the node heuristics to sort the input sequence, while restricting the permutation to a smaller range. 
Specifically, the node degree prioritization setting only allows nodes with the same degree to randomly permute among themselves.

The permutation strategy led to a significant performance gain. The node-level permutation setting saw a 3\% increase in average precision on the Peptides-Func dataset and a 10\% increase in F1 scores on the PascalVOC-SP dataset, compared to the baseline training strategy. 
Combining node prioritization by degree with permutation further improved the scores on the PascalVOC-SP dataset. 
These results underscore the importance of devising graph-focused adaptations of sequence modeling methods, tailoring towards the characteristics of non-sequential data. These elegant sequence design techniques offer improved modeling power to Mamba well beyond simple plug-and-play. We therefore recommend using the combination of input node prioritization by node degree and node-level permutation as the default training and inference recipe for Graph-Mamba. Table \ref{tab:lrgb} and \ref{tab:gnnbenchmark} showcased model performance from this recommended recipe.

\section{Conclusion}
We propose Graph-Mamba, a novel graph model that leverages SSM for efficient data-dependent context selection, as an alternative to graph attention sparsification. GMB's selection mechanism filters relevant nodes as context, effectively compressing and propagating long-range dependencies during node embedding updates. Through recurrent scan with context compression, Graph-Mamba achieves linear-time complexity and reduced memory consumption. The specialized training and inference recipe, combining permutation and node prioritization, further adapts Mamba, a sequence modeling technique, to non-sequential graph input with significant performance improvement. In empirical experiments, Graph-Mamba demonstrated the state-of-the-art or comparable performance across ten datasets with various graph prediction tasks. Graph-Mamba thus presents a promising option to replace traditional dense or sparse graph attention, offering competitive predictive power and long-range context awareness at a fraction of the computational cost. 




In future works, exploring alternative model architectures beyond the GraphGPS framework is crucial for enhancing predictive performance. 
The architecture-agnostic nature of GMB offers flexibility for such applications. Furthermore, this study highlights the significance of sequence construction and training strategies in facilitating sequence model learning with non-sequential graph input. Beyond permutation and node heuristics, effective ways to inject graph topology into input sequences remain unexplored. Ultimately, learning the optimal strategy of flattening a graph into sequences from data is essential. 
SSM-based sequence modeling offers new perspectives for causality analysis beyond prediction, presenting a promising direction for graph data analysis. The improved efficiency further supports the development of graph foundation models, opening up the possibility of large-scale pre-training.

\section*{Impact Statement}
Graph Mamba was designed as a general graph representation learning method. Therefore, it can not have any direct negative societal outcomes. Nonetheless, adverse or malicious application of the proposed algorithm in various domains, including drug discovery and healthcare, may lead to undesirable effects.

\bibliography{main}
\bibliographystyle{icml2024}

\newpage
\appendix
\onecolumn
\section{Dataset Description.} \label{app:dataset}

We evaluate Graph-Mamba on ten datasets from two popular graph benchmarks, the Long-Range Graph Benchmark (LRGB) \cite{dwivedi2023benchmarking} and GNN Benchmark \cite{dwivedi2022long}. Table \ref{tab:dataset} summarizes the dataset characteristics and associated prediction tasks. The first five datasets in bold feature long input size (i.e., Avg. Nodes), corresponding to Table \ref{tab:lrgb}. The other five datasets have small to medium input size, corresponding to Table \ref{tab:gnnbenchmark}.

\textbf{CIFAR10 and MNIST} \cite{dwivedi2023benchmarking} introduce the graph equivalents of the image classification datasets. Each image is represented by the 8-nearest neighbor graph derived from the SLIC superpixels. Both datasets feature 10 classification labels and adhere to the standard dataset splits established by the original image datasets.

\textbf{MalNet-Tiny} \cite{freitas2021largescale} consists of function call graphs extracted from Android APKs. It is a subset of the larger MalNet collection, containing 5,000 graphs each with a maximum of 5,000 nodes. This dataset contains 5 classification labels, including 1 benign software and 4 types of malware. In this benchmarking version, the original node and edge features are removed, and each node is represented with its local degree profile instead. This modification presents a challenging classification task relying solely on graph structures.

\textbf{PATTERN and CLUSTER} \cite{dwivedi2023benchmarking} are synthetic graph datasets of community structures simulated by the Stochastic Block Model (SBM). Both datasets represent node-level classification tasks with an inductive focus. In PATTERN, the goal is to distinguish nodes belonging to 100 distinct sub-graph patterns randomly generated with different SBM parameters from the remaining nodes. In CLUSTER, each graph contains 6 clusters from the same distribution, with 6 test nodes representing each unique cluster. The objective is to predict the cluster identity of these test nodes.


\textbf{Peptides-func and Peptides-struct} \cite{dwivedi2022long} are graph representations of peptides with large diameters. Peptides-Func involves graph-level classification with 10 functional labels. Peptides-Struct focuses on a graph-level regression task, predicting 11 structural properties of the molecules.


\textbf{PCQM-Contact} \cite{dwivedi2022long} introduces a link prediction task based on the PCQM4Mv2 dataset of 3D molecular structures. A contact link is defined between pairs of distant nodes that are spatially close in three-dimensional space. The evaluation metric used is the Mean Reciprocal Rank (MRR), a ranking-based measure.


\textbf{PascalVOC-SP and COCO-SP} \cite{dwivedi2022long} are the graph representations of the image datasets upon SLIC superpixelization. The node-level classification task involves classifying superpixels into corresponding object classes similar to semantic segmentation.


\begin{table*}[h!]
\caption{Dataset description.} 
\centering
 \begin{tabular}{l | r r c c c cccl } 
 \hline \hline
 \textbf{Dataset} & \textbf{Prediction Task} & \textbf{Prediction Level} & \textbf{Graphs} & \textbf{Avg. Nodes} & \textbf{Avg. Edges} & \textbf{Benchmark} \\
\hline
\textbf{Peptides-func}  & Classification & Graph &  15,535 & 150.9 & 307.3 & LRGB \\ 
\textbf{Peptides-struct} & Regression & Graph &  15,535 & 150.9 & 307.3 & LRGB \\ 
\textbf{PascalVOC-SP} & Classification & Node &  11,355 & 479.4 & 2,710.5 & LRGB \\ 
\textbf{COCO-SP}  & Classification & Node &  123,286 & 476.9 & 2,693.7 & LRGB \\ 
\textbf{MalNet-Tiny}  & Classification & Graph &  5,000 & 1,410.3 & 2,859.9 & GNN \\
\hline
PCQM-Contact  & Link Ranking & Link & 529,434 & 30.1 & 61.0 & LRGB \\ 
CIFAR10
  & Classification & Graph & 60,000 & 117.6 & 941.1 & GNN \\
MNIST & Classification & Graph & 70,000 & 70.6 & 564.5 & GNN \\
CLUSTER  & Classification & Node & 12,000 & 117.2 & 2,150.9 & GNN \\ 
PATTERN  & Classification & Node & 14,000 & 118.9 & 3,039.3 & GNN \\
 \hline \hline
\end{tabular}
\vspace{0.1cm}
\label{tab:dataset}
\end{table*}

\section{Additional Benchmark Results.} \label{app:benchmark_supp}

Table \ref{tab:gnnbenchmark} presents benchmark results on the five datasets with small to medium input length, ranging from 30 to 120 nodes in the graph. Graph-Mamba demonstrates predictive performance comparable to GraphGPS with full Transformer and Exphormer, endorsing its generalizability to common graph tasks.

\begin{table*}[t]
\caption{\textbf{Benchmark on Short to Medium-Range Graph Datasets} with existing methods. Best results are colored in \ctext{ForestGreen}{\textbf{first}}, \ctext{BurntOrange}{\textbf{second}}, \ctext{Periwinkle}{\textbf{third}}.} 
\centering
 \begin{tabular}{l c c c c c cccl } 
 \hline \hline
 \textbf{Model} & \textbf{CIFAR10} & \textbf{MNIST}  & \textbf{CLUSTER} & \textbf{PATTERN} & \textbf{PCQM-Contact} \\
  & Accuracy $\uparrow$ & Accuracy $\uparrow$ & Accuracy $\uparrow$ & Accuracy $\uparrow$ & MRR $\uparrow$  \\
 \hline
GCN  & 0.5571$\pm$0.0038 & 0.9071$\pm$0.0021 &  0.6850$\pm$0.0097 & 0.7189$\pm$0.0033 & 0.3234$\pm$0.0006\\
GIN  & 0.5526$\pm$0.0152 & 0.9649$\pm$0.0025 & 0.6472$\pm$0.0155 &  0.8539$\pm$0.0013 & 0.3180$\pm$0.0027\\
GatedGCN  & 0.6731$\pm$0.0031 & 0.9734$\pm$0.0014 &   0.7384$\pm$0.0032 & 0.8557$\pm$0.0008 & 0.3218$\pm$0.0011\\
\hline
GPS+Transformer  & \ctext{Periwinkle}{\textbf{0.7226$\pm$0.0031}} & 0.9811$\pm$0.0011 & \ctext{Periwinkle}{\textbf{0.7799$\pm$0.0017}} & \ctext{BurntOrange}{\textbf{0.8664$\pm$
0.0011}} & \ctext{BurntOrange}{\textbf{0.3442$\pm$
0.0009}} \\
GPS+Performer  & 0.7067$\pm$0.0033 & \ctext{BurntOrange}{\textbf{0.9834$\pm$0.0003}}  & \ctext{ForestGreen}{\textbf{0.7829$\pm$0.0004}} & 0.8334$\pm$0.0029 & \ctext{Periwinkle}{\textbf{0.3437$\pm$0.0005}} \\
GPS+BigBird  & 0.7048$\pm$0.0010 & 0.9817$\pm$0.0001 & 0.7746$\pm$0.0002 & 0.8600$\pm$0.0014 & 0.3391$\pm$0.0002 \\
Exphormer  & \ctext{ForestGreen}{\textbf{0.7413$\pm$0.0050}} & \ctext{ForestGreen}{\textbf{0.9843$\pm$0.0004}} & \ctext{BurntOrange}{\textbf{0.7802$\pm$0.0011}} & \ctext{ForestGreen}{\textbf{0.8670$\pm$0.0003}} & \ctext{ForestGreen}{\textbf{0.3587$\pm$
0.0025}} \\
\hline
Graph-Mamba  & \ctext{BurntOrange}{\textbf{0.7370$\pm$0.0034}} & \ctext{ForestGreen}{\textbf{0.9842$\pm$0.0008}} & 0.7680$\pm$0.0036 &  \ctext{ForestGreen}{\textbf{0.8671$\pm$0.0005}} & 0.3395$\pm$
0.0013\\ 
 \hline \hline
\end{tabular}
\vspace{0.1cm}
\label{tab:gnnbenchmark}
\end{table*}

\section{Proof of Theorem.}\label{proof-theorem}
We revisit the proof of the main Theorem in Mamba as presented by \citet{gu2023mamba} for further reference for the SSM selection mechanism.

\begin{namedtheorem}
\textit{Consider N = 1, $\bm{A}$ = -1, $\bm{B}$ = 1, and $\Delta_{t} = softplus(Linear(x_{t}))$ for selective SSM, the discretized recurrence output is defined as} 
\begin{equation} \label{proof}
\begin{split}
    g_{t} =& \sigma(Linear(x_{t}))\\
    h_{t} =& (1-g_{t})h_{t-1} + g_{t}x_{t}
\end{split}
\end{equation}
\end{namedtheorem}

\begin{proof}
Substitute the expression of $\Delta_{t}$ into the zero-order hold discretization formulas $\bm{\bar{A}}=\exp(\Delta_{t} \bm{A})$ and $\bm{\bar{B}}=(\Delta_{t} \bm{A})^{-1}(\exp(\Delta_{t}\cdot\bm{A})-I)\cdot \Delta_{t} \bm{B}$.
\begin{equation}
\begin{split}
    \bm{\bar{A}_{t}} 
    &= exp(\Delta_{t}\bm{A}) \\
    &= exp(- softplus(Linear(x_{t}))) \\
    &= \frac{1}{1+exp(Linear(x_{t}))}\\
    &= \sigma(-Linear(x_{t}))\\
    &= 1- \sigma(Linear(x_{t}))\\
    \bm{\bar{B_{t}}} & = (\Delta_{t} \bm{A})^{-1}(exp(\Delta_{t} \bm{A}) - \bm{I}) \cdot \Delta_{t} \bm{B}\\
    &= -(exp(\Delta_{t} \bm{A}) - \bm{I})\\
    &= 1-\bm{\bar{A_{t}}}\\
    &= \sigma(Linear(x_{t}))\\
\end{split}
\end{equation}
\end{proof}

\section{Input Node Prioritization and Permutation Strategies.} \label{supp:seqconstruction}

To elaborate on the rationale behind node prioritization, we first assume a sequence of $L$ nodes in random order, denoted as $ND_{0}$, $ND_{1}$, ... , $ND_{L-1}$. The selection mechanism implemented as SSM computation in the Mamba module performs a unidirectional scan over the sequence of nodes, updating the hidden states at each step of recurrence. This defines different levels of access to context (i.e., other nodes) based on node position in the sequence. For example, $ND_{1}$ has limited access to context and updates itself based on hidden states that encode $ND_{0}$ only. Its connections to the other nodes are removed entirely. In contrast, $ND_{L-1}$ has access to most context including all prior nodes $ND_{0}$ to $ND_{L-2}$. It reatins connections to all other nodes. Intuitively, with a randomly ordered node sequence, SSM creates a similar effect as random subsampling. Instead, we would like to create a biased sampling procedure that favor connections for important nodes in a graph. This is achieved by placing the important nodes at the end of the sequence.

In addition to the recommended recipe, we explored a few variants of node prioritization and permutation techniques summarized as follows:

\begin{itemize}
    \item \textbf{Node prioritization by eigenvector centrality.} We explored eigenvector centrality as an alternative proxy for node importance. Eigenvector centrality measures a node's influence by adding the centrality of its neighbors. A node with high centrality indicates connections to other influential nodes in the graph. The eigenvalue centrality of each node is defined by the principal eigenvector that corresponds to the largest eigenvalue. In Graph-Mamba, the eigenvector centrality scores were calculated using the \texttt{eigenvector\_centrality\_numpy()} function from \texttt{NetworkX}.
    \item \textbf{Permutation by clusters.}  The input sequence consists of nodes grouped by clusters defined by edge connectivity. The permutation happens first on the cluster level. The nodes within each cluster are then randomly permuted among themselves. The intuition behind is that the local structure and topology are essential and not represented in dense attention. This approach aims to inject the local structure into the input sequence. We used the Louvain algorithm for unsupervised graph partitioning.
\end{itemize}

Table \ref{tab:seq_sort_2} presents the predictive performance from these two aforementioned variants, highlighted in blue. Cluster-level permutation leads to a significant performance gain compared to the baseline, but slightly less compared to node-level permutation. Similarly, node prioritization by eigenvector centrality is not as effective as by degree as demonstrated in these two datasets. 

\begin{table*}[h!]
\caption{\textbf{Comparison with Alternative Node Prioritization and Permutation Strategies} on the Peptides-Func and PascalVOC-SP datasets.} 
\centering
 \begin{tabular}{ l | l | c c } 
 \hline \hline
  \textbf{Permutation} & \textbf{Node Prioritization} & \textbf{Peptides-Func} & \textbf{PascalVOC-SP} \\ 
  & & AP $\uparrow$ &  F1 score $\uparrow$ \\
 \hline
 \hline
   - &  - & 0.6581 &  0.3105\\ 
\hline
   Node Level & - & 0.6821 & 0.4193 \\ 
   \textcolor{blue}{Cluster Level} & \textcolor{blue}{-} & \textcolor{blue}{0.6769} & \textcolor{blue}{0.3802} \\ 
\hline
  \textcolor{blue}{Node Level} & \textcolor{blue}{Eigenvector Centrality} & \textcolor{blue}{0.6739} & \textcolor{blue}{0.3961}

\\
   Node Level & Degree & \textbf{0.6834} & \textbf{0.4314}
\\
 \hline \hline
\end{tabular}
\vspace{0.1cm}
\label{tab:seq_sort_2}
\end{table*}

\section{Binning Technique for Large Graphs.} \label{app:bin}
 For large graph datasets, we devised a binning technique that randomly divides the long sequence of input nodes into $n$ bins and applies GMB to each of the sub-sequence individually. Node prioritization and permutation happen within each sub-sequence. We then obtain the GMB output for each sub-sequence in $n$ passes, and combine the updated sub-sequences to match the node order in the original sequence. Conceptually, the binning technique further sparsifies the node connections for embedding updates, since only nodes within a sub-sequence can interact. This technique helps further reduce memory usage.
 
\section{Hyperparameters.} \label{app:hp}

We followed the hyperparamter suggestions in the Exphormer benchmark \cite{shirzad2023exphormer}. We matched the number of parameters in the Mamba/attention module and the whole model between Graph-Mamba and GPS+Transformer. The model size, positional encodings, and batch size remained consistent with the default in the Exphormer benchmark across all models. For Graph-Mamba, we used a Mamba block with state dimension of 16, convolution kernel size of 4, and expansion factor of 1. We adjusted the Adam optimizer setting by increasing the initial learning rate and introducing weight decay, following Mamba's training recipe \cite{gu2023mamba}.

Tables \ref{tab:hyperparamLRGB} and \ref{tab:hyperparamGNNBenchmark} summarize the hyperparameters used for Graph-Mamba training. For trainable parameters, ``Num Params Mamba'' indicates the number of parameters in a single Mamba block in one GMB layer, and ``Num Params Total'' reports the total number of trainable parameters in the model. The binning technique described in Appendix \ref{app:bin} is only applied to large graph datasets with close to or more than 500 nodes.

\begin{table*}[h!]
\caption{\textbf{Hyperparameters used for Graph-Mamba on Long-Range Graph Datasets.}} 
\centering
 \begin{tabular}{l | c c c c c ccccccccl } 
 \hline \hline
 \textbf{Dataset} & \textbf{Peptides-Func} & \textbf{Peptides-Struct} & \textbf{PascalVOC-SP} & \textbf{COCO-SP} & \textbf{MalNet-Tiny} \\
 \hline
Num Layers & 4 & 4 & 4 & 4 & 5\\
Hidden Dim & 96 & 96 & 96 & 96 & 64\\
PE  & LapPE & LapPE & LapPE & LapPE & LapPE \\ 
\hline
Batch Size  & 128 & 128 & 32 & 32 & 16 \\
Learning Rate (LR)  & 0.001 & 0.001 & 0.0015 & 0.0015 & 0.0015 \\ 
Weight Decay & 0.01 & 0.01 & 0.001 & 0.001 & 0.00001 \\ 
Num Epochs  & 200 & 200 & 300 & 300 &  150 \\ 
\hline
Num Bins & - & - & 2 & 2 & 3  \\
\hline
Num Params Mamba & 34,080 & 34,080 &  34,080 & 34,080 & 16,320 \\ 
Num Params Total  & 373,018 & 491,867 &  497,781 & 497,781 & 285,125 \\ 
 \hline \hline
\end{tabular}
\vspace{0.1cm}
\label{tab:hyperparamLRGB}
\end{table*}

\begin{table*}[h!]
\caption{\textbf{Hyperparameters used for Graph-Mamba on Short to Medium-Range Graph Datasets.}} 
\centering
 \begin{tabular}{l | c c c c c ccccccccl } 
 \hline \hline
 \textbf{Dataset} & \textbf{CIFAR10} & \textbf{MNIST} & \textbf{CLUSTER} & \textbf{PATTERN} & \textbf{PCQM-Contact} \\
 \hline
Num Layers & 3 & 3 & 16 & 6 & 4 \\
Hidden Dim & 52 & 52 & 48 & 64 & 96 \\
PE  & LapPE & LapPE & LapPE & LapPE & LapPE \\ 
\hline
Batch Size  & 16 & 16 & 16 & 32 & 128\\
Learning Rate (LR)  & 0.005  & 0.005 & 0.001 & 0.001 & 0.002\\ 
LR Weight Decay & 0.01 & 0.01 & 0.0001 & 0.0001 & 0.01  \\ 
Num Epochs  & 100 & 100 & 100 & 100 & 100 \\ 
\hline
Num Bins & - & - & - & - &  - \\
\hline
Num Params Mamba & 11,388 & 11,388 & 9,840 & 16,320 & 34,080\\ 
Num Params Total  & 113,818 & 116,486  & 508,966 & 335,281 & 500,112\\ 
 \hline \hline
\end{tabular}
\vspace{0.1cm}
\label{tab:hyperparamGNNBenchmark}
\end{table*}

\section{Benchmarking Experiments.} \label{app:exp}

For the GNN and LRGB benchmarks, we reported the average over 5 runs of random seeds 0-4 for Graph-Mamba, GPS+Transformer, and Exphormer. For the earlier methods and some of the GPS+Performer and GPS+BigBird runs, we consolidated the scores from the Exphormer benchmark \cite{shirzad2023exphormer}. We reported the standard evaluation metrics for each dataset using the same pipelines as GraphGPS and Exphormer. Note that we capped the maximum GPU memory at 24GB for this benchmark, and an OOM case was reported when reducing the standard batch size by half still led to an OOM error. For the ablation study on training and inference recipes, we reported the scores from a single run with random seed 0.

For the FLOPs and memory benchmark, the statistics were collected from a single epoch in training phase. Specifically, we reported the average FLOPs performed per sample by summing the total number of FLOPs performed in all forward passes in one epoch, and dividing it by the number of training examples. Similarly for memory, we reported the peak GPU memory usage divided by the batch size as an estimate for average memory usage per training example. For FLOPs profiling experiments, we used the \texttt{FlopsProfiler} implementation from the \texttt{deepspeed} package. We obtained the peak GPU memory usage in one epoch from the \texttt{torch.cuda.max\_memory\_allocated()} function. The MalNet-Tiny dataset features the largest input graphs with an average of 1,410.3 nodes. To assess how the models scale with input size, we subsampled the Malnet-Tiny dataset at various ratios including 0.05, 0.1, 0.2, 0.3, 0.4, 0.5, 0.6, 0.7, 0.8, and 0.9 to simulate increasing number of nodes in the input. Specifically, for each input graph, we randomly selected a fraction of input nodes at the desired ratio, and retained the edges associated with these selected nodes only. To ensure a fair comparison, we matched the model size for GPS+Transformer (286,725 parameters) and Graph-Mamba (285,125 parameters), consistent with the predictive performance benchmark. For the subsequent benchmark including the GPS sparse attention variants, the original datasets PCQM-Contact, CLUSTER, Peptides-Func, and PascalVOC-SP were used without any subsampling. These datasets feature increasing input length ranging from 30 up to 480 nodes.

\section{Implementation Details.}
 The Graph-Mamba model is implemented in \texttt{PyTorch} framework using \texttt{mamba-ssm} and \texttt{PyTorch Geometric}. The model was trained on a single \texttt{RTX6000} or \texttt{A100} GPU. 


\end{document}